\documentclass[letterpaper, 10 pt, conference]{ieeeconf}  
\let\labelindent\relax
\usepackage{graphicx} 
\usepackage{multirow} 
\usepackage{enumitem} 
\usepackage[margin=1in]{geometry}
\usepackage{caption}
\usepackage{float}
\usepackage{amsmath}
\usepackage{url}

\graphicspath{{fig/}}
\DeclareUnicodeCharacter{202F}{\,}

\IEEEoverridecommandlockouts                              

\usepackage{amssymb}  
\usepackage{multirow}

\usepackage{marvosym}
\usepackage{makecell}

\usepackage{graphics} 
\usepackage{epsfig} 
\usepackage{times} 
\usepackage{stfloats}
\usepackage{enumerate}

\title{\LARGE \bf
Bench2FreeAD: A Benchmark for End-to-end Navigation in Unstructured Robotic Environments 
}
\author{Yuhang Peng, Sidong Wang, Jihaoyu Yang, Shilong Li, Han Wang, and Jiangtao Gong%
\thanks{Corresponding author Jiangtao Gong (gongjiangtao@air.tsinghua.edu.cn)}
}

\begin{document}

\maketitle
\thispagestyle{empty}
\pagestyle{empty}

\begin{abstract}


Most current end-to-end (E2E) autonomous driving algorithms are built on standard vehicles in structured transportation scenarios, lacking exploration of robot navigation for unstructured scenarios such as auxiliary roads, campus roads, and indoor settings. This paper investigates E2E robot navigation in unstructured road environments. First, we present two data collection pipelines: one for acquiring real-world robotic data, and another for generating synthetic data using a Unity-based simulator. which together produce an unstructured robotics navigation dataset---FreeWorld Dataset. Second, we fine-tuned two efficient E2E autonomous driving model, VAD and LAW, using our data sets to validate the performance and adaptability of E2E autonomous driving models in these environments. Results demonstrate that fine-tuning through our dataset significantly enhances the navigation potential of E2E autonomous driving models in unstructured robotic environments. Thus, this paper presents the first dataset targeting E2E robot navigation tasks in unstructured scenarios, and provides a benchmark based on vision-based E2E autonomous driving algorithms to facilitate the development of E2E navigation technology for logistics and service robots. The project is available on Github\footnote{\url{https://github.com/AIR-DISCOVER/FreeAD/}}.

\end{abstract}

\begin{figure}[htp]
    \centering
    \includegraphics[width=7cm]{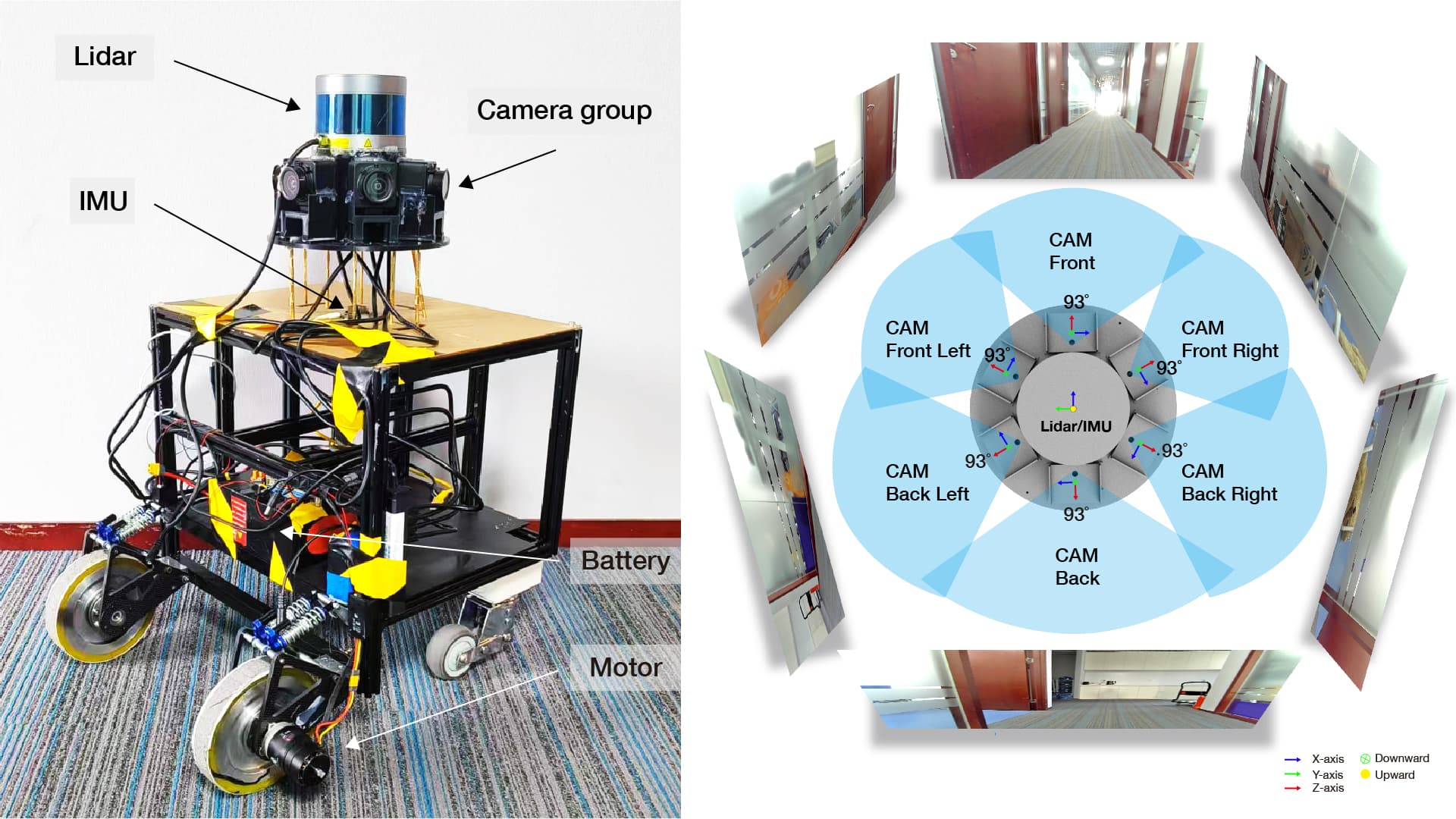}
    \caption{Overview of the real vehicle setup, sensor coordinate system, and camera group field of view.}
    \label{fig:overview_of_robot_compressed}
\end{figure}

\section{INTRODUCTION}
In recent years, end-to-end (E2E) autonomous driving models have made remarkable progress, revolutionizing the field of autonomous transportation. Models such as DETR3D~\cite{wang2022detr3d}, BEVFormer~\cite{li2024bevformer}, and UniAD~\cite{hu2023planning} have demonstrated impressive capabilities in structured environments by directly mapping sensory inputs to driving actions without explicit intermediate steps. These approaches have shown particular promise in urban settings with well-defined road structures, traffic rules, and predictable agent behaviors. The VAD model~\cite{jiang2023vad} further advanced this field by utilizing vectorized bird's-eye view (BEV) representations, enabling more efficient real-time decision-making while maintaining high performance standards. Despite these advancements, most E2E driving models have been primarily developed and evaluated on datasets like nuScenes~\cite{caesar2020nuscenes}, which focus on standard high-speed vehicles in structured urban environments.

Concurrently, robot navigation technologies have evolved along several paths. Traditional approaches relied on rule-based methods such as A* and Dijkstra's algorithm~\cite{xiao2022motion}, which perform well in controlled environments but struggle with uncertainty and dynamic scenarios. Statistical methods, including SLAM techniques~\cite{macenski2021slam}, improved adaptability by incorporating probabilistic frameworks that better handle sensor noise and partial observability. More recently, learning-based methods have gained prominence, with Deep Reinforcement Learning (DRL)\cite{zhu2021deep} enabling robots to learn complex behaviors through environmental interaction. Imitation learning approaches\cite{hussein2017imitation} have shown particular promise by allowing robots to mimic expert demonstrations, accelerating the acquisition of effective navigation strategies without extensive trial-and-error.

Despite these parallel advancements, autonomous navigation in unstructured environments presents persistent challenges that remain inadequately addressed. These environments—including auxiliary roads, low-speed zones, indoor spaces, and irregularly shaped pathways—lack the clear demarcations, consistent surfaces, and predictable layouts of structured settings. Navigation in such contexts requires robust handling of ambiguous boundaries, variable terrain, unexpected obstacles, and complex interactions with pedestrians and other dynamic agents. Current datasets like Replica~\cite{straub2019replica}, Matterport3D~\cite{chang2017matterport3d}, and GOOSE~\cite{mortimer2024goose} provide valuable resources but lack comprehensive coverage of unstructured scenarios with appropriate annotations for end-to-end autonomous navigation.


In this paper, we address these challenges by developing a comprehensive approach to E2E robot navigation in unstructured environments. We introduce two complementary data collection pipelines: one gathering real-world robot navigation data across various unstructured settings, and another generating synthetic data using the Unity-based simulation environment. These pipelines yield a rich, diverse dataset specifically designed for training and evaluating navigation models in unstructured scenarios. Our dataset features annotations of road dividers as the primary static object class (with all other areas considered navigable) and 3D bounding boxes for dynamic objects, particularly pedestrians and cars.

Building on this dataset, we fine-tune the VAD~\cite{jiang2023vad} and LAW~\cite{li2024enhancing} model for unstructured environment navigation. This adaptation process involves adjusting the model architecture to better handle the unique characteristics of unstructured environments while maintaining computational efficiency for deployment on resource-constrained robot platforms. Our fine-tuning methodology balances performance enhancement with generalization capabilities, ensuring the model functions effectively across a range of previously unseen unstructured scenarios.

The contributions of this work are threefold:
\begin{itemize}
    \item We present the first comprehensive dataset specifically designed for E2E robot navigation in unstructured environments, featuring both real-world and synthetic data with appropriate annotations for static maps and dynamic objects.
    \item We demonstrate the successful adaptation of the VAD and LAW model to unstructured navigation contexts through a specialized fine-tuning process, significantly enhancing its performance in these challenging scenarios.
    \item We establish a benchmark for vision-based E2E autonomous navigation in unstructured environments, providing a foundation for future research in this emerging field and facilitating the development of navigation technologies for logistics and service robots.
\end{itemize}




\begin{figure}[htp]
    \centering
    \includegraphics[width=7.8cm]{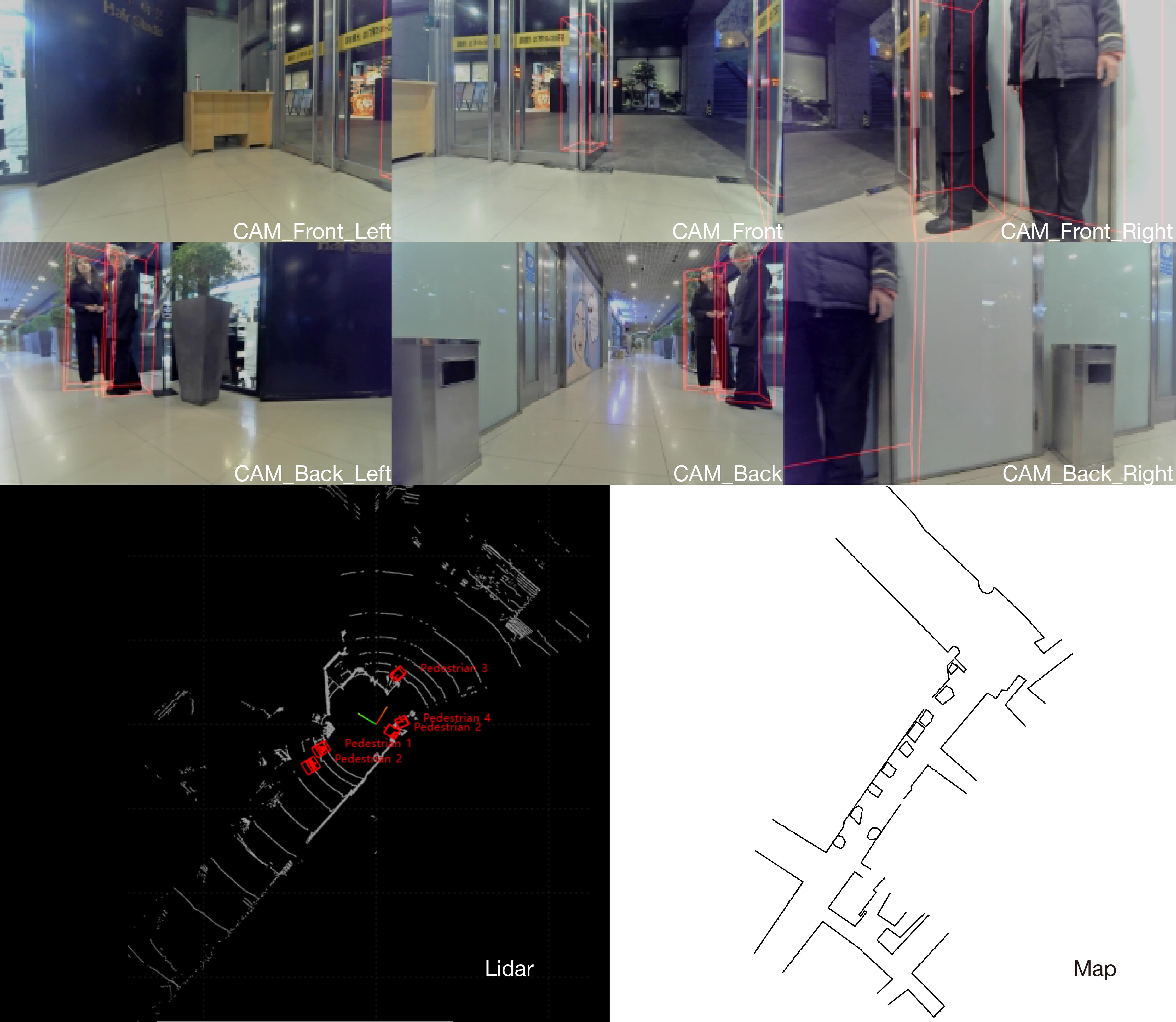}
    \caption{An example from the FreeWorld dataset. We see 6 different camera views and lidar data, as well as the human annotated semantic map. At the bottom we show the human written scene description.}
    \label{fig:6cam_Lidar_Map_compressed}
\end{figure}

\begin{table*}[htbp]
\caption{Comparison of Different Datasets. 
We extracted dataset statistics primarily from the respective publications. 
If essential details were missing and could not be confirmed via author communication, we mark them as \textit{N/A}. 
If a dataset inherently does not provide a certain type of information, we denote it with ``--''. 
Pose: G = GPS/GNSS; R = relative odometry (IMU/LiDAR/visual-inertial); S = SLAM-estimated trajectory.}
\label{tab:ComparisionOfDifferentDatasets}
\centering
\renewcommand{\arraystretch}{1.3} 
\setlength{\tabcolsep}{3pt} 
\begin{tabular}{|l|c|c|c|c|c|c|c|c|c|c|c|}
\hline
\textbf{Dataset} & \textbf{Pose} & \textbf{Clxzs} & \textbf{3D Annotation} & \textbf{2D Frames} & \textbf{Sequences} & \makecell{\textbf{Time}\\\textbf{of Day}} & \textbf{Real/Sim.} & \makecell{\textbf{Environment}} & \makecell{\textbf{Map Info}} & \makecell{\textbf{Map Type}} \\
\hline
KITTI~\cite{geiger2012we}        & G+R  & 3   & 15K   & 13K   & 22    & M, A    & Y/N & Outdoor        & --                  & -- \\
nuScenes~\cite{caesar2020nuscenes} & G+R  & 23  & 40K   & 1.4M  & 1000  & M, A, E & Y/N & Outdoor        & 4 maps   & vector \\
CODa~\cite{li2022coda}           & S+G  & 53  & 32K   & 131K  & 54    & M, A, E & Y/N & Outdoor & --                  & -- \\
JRDB~\cite{le2024jrdb}           & --   & 1   & 28K   & 28K   & --    & M, A    & Y/N & Outdoor/Indoor & --                  & -- \\
SCAND~\cite{karnan2022socially}  & --   & --   & --     & 626K  & --    & M, A    & Y/N & Outdoor/Indoor & --                  & -- \\
TartanAir~\cite{wang2020tartanair} & --   & 6   & N/A   & N/A   & 1037  & M, A    & N/Y & Outdoor/Indoor & --                  & occupancy \\
M2DGR~\cite{yin2021m2dgr}        & G+R  & --  & --   & N/A   & 36    & M, A    & Y/N & Outdoor/Indoor & --                  & -- \\
ViViD++~\cite{lee2022vivid}      & G+R  & --  & --    & --    & 36    & M, A, E & Y/N & Outdoor/Indoor & --                  & -- \\
Hilti-Oxford~\cite{zhang2022hilti} & G+R & --  & --   & N/A   & 16    & -- & Y/N & Indoor         & --                  & -- \\
MCD~\cite{nguyen2024mcd}         & R    & 29  & N/A   & N/A   & 18    & M, A, E & Y/N & Outdoor/Indoor         & --                  & -- \\
MulRan~\cite{kim2020mulran}      & G    & --  & --    & --    & 10    & M, A    & Y/N & Outdoor        & --                  & -- \\
SubT-MRS~\cite{zhao2024subt}     & G    & --  & --    & N/A   & --    & M, A, E & Y/N & Outdoor/Indoor & --                  & -- \\
360Loc~\cite{huang2024360loc}    & R    & --  & --    & 56.72k& 18    & M, A, E & Y/N & Outdoor/Indoor & --                  & -- \\
NCLT~\cite{carlevaris2016university} & G+R & --  & -- & N/A   & 27    & M, A, E & Y/N & Outdoor/Indoor & --                  & occupancy \\
HuRoN~\cite{shah2021rapid}       & G+R  & --  & --    & N/A   & N/A   & -- & Y/N & Indoor         & --                  & topological \\
FreeWorld (Ours)                 & S+R+G  & 16   & 1.5M    & 130K   & 309    & M, A, E    & Y/Y & Outdoor/Indoor & 6 maps     & vector, occupancy \\
\hline
\end{tabular}
\end{table*}

\begin{figure}[htp]
    \centering
    \includegraphics[width=7.8cm]{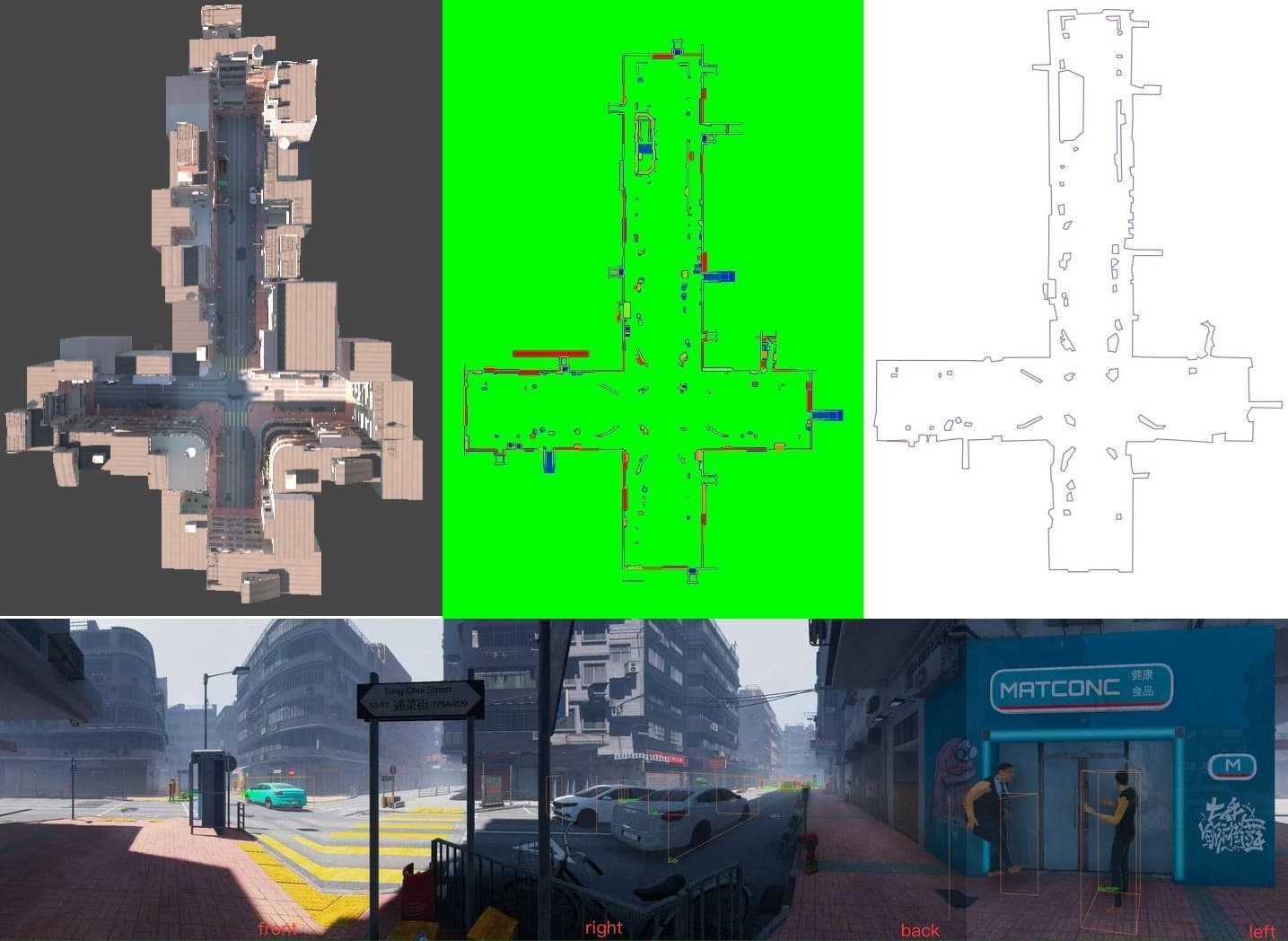}
    \caption{The upper part illustrates the grid-based heat occupancy map and its vectorized representation generated in the Unity simulation environment. The lower part shows 3D bounding boxes of object instances (e.g., humans and cars) produced by the Unity Perception package.}
    \label{fig:HongKongCityOccupancyAndVector}
\end{figure}

\begin{table}[ht]
    \centering
    \begin{tabular}{|c|c|}
        \hline
        \textbf{Sensor Type} & \textbf{Specifications} \\
        \hline
        \multirow{7}{*}{Camera} & 6x Camera RGB \\
        & 7Hz capture frequency \\
        & 1/2.5” IMX274 sensor \\
        & HFOV: 93° \\
        & VFOV: 61° \\
        & 1920 × 1080 resolution, \\&auto exposure, JPEG compressed \\
        \hline
        \multirow{6}{*}{Lidar} & 1x Lidar Spinning, 16 beams \\
        & 20Hz capture frequency \\
        & 360° horizontal FOV \\
        & \textminus 15° to 15° vertical FOV \\
        & $\leq$ 150m range, $\pm$2cm accuracy,\\& up to 300k points per second \\
        \hline
        \multirow{3}{*}{IMU} & IMU, HI229 \\
        & 10° heading, 0.8° roll/pitch \\
        & 400Hz update rate \\
        \hline
    \end{tabular}
    \caption{Sensor specifications in the real world}
    \label{tab:sensor_details}
\end{table}

\section{The FreeWorld Dataset}

\subsection{Dataset Overview}
We collected more than 2,000 samples in 3 hours in real world, covering five maps with vector annotations across 50 scenes, and annotated more than 2,000 3D bounding boxes. An example is shown in Fig.\ref{fig:6cam_Lidar_Map_compressed}. Our dataset consists of two primary modalities: images and LiDAR point clouds. Images are mainly used to train vision-based models, whereas LiDAR point clouds are mainly used for map reconstruction. We further collected 63,429 simulated samples within 17 hours using the Unity-based simulator, which we refer to as the virtual environment dataset. These samples include diverse driving scenarios with vectorized annotations. In addition, we used a rule-based algorithm to generate trajectories, which serve as baseline references to evaluate motion prediction and planning tasks.

\subsection{Dataset Comparison}
By Table~\ref{tab:ComparisionOfDifferentDatasets}, our benchmark distinguishes itself by providing full coverage $360^{\circ}$, while many existing robotic datasets are limited to monocular or single-direction views~\cite{li2022coda,wang2020tartanair,yin2021m2dgr,lee2022vivid,zhang2022hilti,nguyen2024mcd,kim2020mulran,carlevaris2016university,shah2021rapid,zhao2024subt}. This restriction hinders the applicability of BEV-based methods in unstructured environments, where broader scene coverage is essential. In addition, a large portion of robotics datasets lack map annotations, which further limits the ability of downstream methods to generate accurate and reliable BEV representations~\cite{li2022coda,le2024jrdb,karnan2022socially,yin2021m2dgr,lee2022vivid,zhang2022hilti,nguyen2024mcd,kim2020mulran,zhao2024subt,huang2024360loc}.

\subsection{Benchmark}
We use a different benchmark for map definition compared to the nuScenes~\cite{caesar2020nuscenes} dataset, as our environment is unstructured. Instead of following the structured mapping approach used in MapTR~\cite{liao2022maptr}, we model the ground truth (GT) map using a simplified semantic layer that focuses primarily on road dividers. Although we initially experimented with the incorporation of both boundaries and dividers, we found that it was unnecessary. Road dividers serve as representations of boundaries and static obstacles that cannot be traversed. All other areas are considered navigable. We believe that reducing the emphasis on semantic segmentation will help the model focus more on obstacle detection and learn their general characteristics.

\subsection{Robot Setup}
\subsubsection{Real World}
We employ a custom-built small robot, with the sensor layout illustrated in Fig.~\ref{fig:overview_of_robot_compressed} and detailed specifications provided in Table \ref{tab:sensor_details}. Unlike nuScenes, our LiDAR coordinates align with the ego frame. The sensors are securely mounted using a custom designed module (shown in Fig.~\ref{fig:overview_of_robot_compressed}) to ensure stable and precise positioning. Each of the six cameras is equipped with a horizontal field of view (HFOV) of 93 ° and a vertical field of view (VFOV) of 61 °. This wide field of view allows the robot to capture a more comprehensive perspective at close distances, ensuring adequate overlap between the views of different cameras, as shown in Fig.~\ref{fig:overview_of_robot_compressed}. In contrast to outdoor vehicles, which typically possess longer detection ranges, the cameras in our robot are specifically optimized for close-range, high-resolution observation.

\subsubsection{Virtual Environment}
The virtual environment is constructed using the Unity High-Definition Render Pipeline (HDRP). A sensor group is mounted at a human-like height of 1.6 meters, serving as the reference point for data acquisition. All six cameras are co-located at this position but oriented in different directions, thereby eliminating the need for explicit extrinsic calibration. Together, the six cameras provide a seamless panoramic viewpoint for comprehensive scene perception.

\subsection{Sensor Synchronization}
In our real-world robotic setup, we employ the ROS2 approximate-time synchronizer to align sensor data. Measurements are filtered if the time difference between the main sensors (six cameras and one LiDAR) exceeds 0.2 s. When a main sensor sample is captured, the IMU and odometry measurements are subsequently triggered to ensure temporal synchronization. In the Unity simulation environment, sensor synchronization is achieved by triggering all virtual sensors within the same rendering frame, ensuring that their outputs are temporally aligned.

\subsection{Imitation Behavior}
In the real-world setup, the robot is teleoperated via remote control, enabling the operator to navigate toward designated targets while avoiding dynamic obstacles such as pedestrians. This procedure yields valuable datasets that capture human-like navigation behaviors. In the virtual environment, Unity’s navigation mesh (NavMesh) framework is employed in combination with an A* algorithm for global trajectory planning and the Social Force Model (SFM) for local obstacle avoidance. Together, these components generate high-quality expert demonstrations of navigation strategies, which serve as a foundation for subsequent learning and evaluation.

\subsection{Localization}
In the real-world setup, localization is achieved using a LiDAR-SLAM algorithm, as GPS signals are unavailable in indoor environments. This method demonstrates high robustness, consistently maintaining localization errors within 10 cm during data collection. In contrast, within the Unity-based virtual environment, precise positional information is directly obtained from the Unity system or its Perception~\cite{borkman2021unity} package, providing global or camera pixel coordinate data without the need for additional localization algorithms.

\subsection{Mapping}
In the real-world deployment, a LiDAR-SLAM algorithm was employed to perform mapping across both indoor and outdoor environments. In Unity-based simulation, we developed a custom voxel-based method to generate a grid-structured occupancy map. This approach produces a heat-based spatial representation of the environment, as illustrated in Fig.~\ref{fig:HongKongCityOccupancyAndVector}.

\subsection{Data Annotation}
In the real-world scenario, 3D bounding box annotations are generated using the joint annotation tool described in \cite{li2020sustech}, which integrates LiDAR and image data to improve the efficiency and precision of labeling. In the Unity simulation, we leverage the Perception package \cite{borkman2021unity} to automatically generate 3D bounding boxes, as well as additional supervisory signals such as semantic segmentation images and related annotations.

\begin{figure*}[t]
    \centering
    \includegraphics[width=0.8\textwidth]{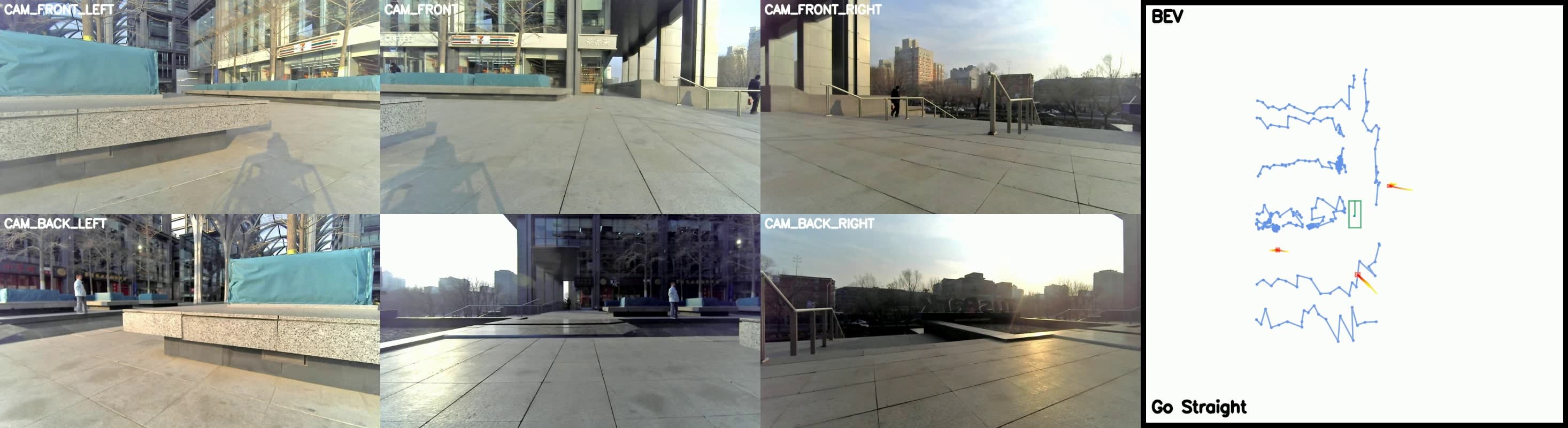}
    \caption{Qualitative results of VAD(r). VAD(r) generates vectorized representations of unstructured scenes and predicts 3D bounding boxes for people.}
    \label{fig:qualitative_analysis_compressed}
\end{figure*}

\section{Metrics}
\subsection{Average Precision}
We adopt the Average Precision (AP) metric with a twist: matches are based on the 2D center distance \(d\) on the ground plane instead of IoU. This decouples detection accuracy from object size and orientation, making comparisons with vision-only methods fairer. AP is computed as the normalized area under the precision-recall curve, considering only operating points where both recall and precision exceed 10\%. If none exist, AP is zero.

\subsection{L2 Error}
The L2 error is defined as the mean Euclidean distance between predicted and ground truth trajectories over \(T\) steps, i.e., \(\text{L2 Error} = \frac{1}{T}\sum_{t=1}^{T}\|\tau_t - \tau^{\mathrm{gt}}_t\|_2\).

\subsection{Collision Rate}
Collision rate quantifies how often the planned trajectory intersects obstacles. At each step, the vehicle's polygonal footprint is projected onto the BEV map. A collision is flagged if any pixel overlaps an obstacle. For \(N\) trajectories, it is defined as \(\text{Collision Rate} = \frac{1}{N}\sum_{i=1}^{N}\sum_{t=1}^{T}\mathbb{1}(\text{collision}_{i,t})\), where \(\mathbb{1}(\cdot)=1\) if a collision occurs, and 0 otherwise.

\begin{table*}[htbp]
\centering
\caption{Performance comparison of open-loop L2 error and collision rate for different fine-tuned models on the FreeWorld dataset in virtual or real part}
\resizebox{\textwidth}{!}{%
\begin{tabular}{llcccccccc}
\hline
\textbf{Training Data} & \textbf{Testing Data} & \multicolumn{4}{c}{\textbf{L2 (m) }$\downarrow$} & \multicolumn{4}{c}{\textbf{Collision (\%) }$\downarrow$} \\
\cline{3-10}
 &  & \textbf{1s} & \textbf{2s} & \textbf{3s} & \textbf{Avg.} & \textbf{1s} & \textbf{2s} & \textbf{3s} & \textbf{Avg.} \\
\hline
VAD          & Virtual & 8.38 & 13.66 & 18.61 & 13.55 & 0.0462 & 0.0231 & 0.0154 & 0.0282 \\
VAD(v)     & Virtual & 8.03 & 13.02 & 17.65 & 12.90 & 0.0465 & 0.0232 & 0.0155 & 0.0284 \\
VAD(r)       & Virtual & 4.13 & 6.33  & 7.87  & 6.11  & 0.0405 & 0.0202 & 0.0135 & 0.0247 \\
VAD(v+r)     & Virtual & 3.74 & 6.14  & 8.24  & 6.04  & 0.0418 & 0.0209 & 0.0139 & 0.0255 \\
LAW          & Virtual & 7.29 & 13.70 & 20.28 & 13.75 & 0.0440 & 0.0220 & 0.0150 & 0.0270 \\
LAW(v)       & Virtual & \textbf{2.16} & \textbf{3.30} & \textbf{3.75}  & \textbf{3.07}  & 0.0218 & 0.0109 & 0.0073 & 0.0133 \\
LAW(r)       & Virtual & 2.32 & 3.81 & 4.80  & 3.64  & 0.0230 & 0.0110 & 0.0080 & 0.0140 \\
LAW(v+r)     & Virtual & 2.20 & 3.39 & 3.99  & 3.19  & \textbf{0.0206} & \textbf{0.0103} & \textbf{0.0069} & \textbf{0.0126} \\
\hline
VAD(v)       & Real    & 0.42  & 0.59  & 0.75  & 0.59  & 0.0066 & 0.0166 & 0.0159 & 0.0130 \\
VAD(r)       & Real    & 0.43  & 0.61  & 0.77  & 0.60  & 0.0070 & 0.0077 & 0.0100 & 0.0082 \\
VAD(v+r)     & Real    & 0.43  & 0.61  & 0.78  & 0.61  & 0.0151 & 0.0219 & 0.0250 & 0.0207 \\
LAW          & Real    & 13.21 & 22.08 & 30.91 & 22.06 & \textbf{0.0009} & \textbf{0.0005} & \textbf{0.0003} & \textbf{0.0006} \\
LAW(v)       & Real    & 0.28  & 0.43  & 0.58  & 0.43  & 0.0027 & 0.0045 & 0.0102 & 0.0058 \\
LAW(r)       & Real    & 0.28  & 0.43  & 0.57  & 0.42  & \textbf{0.0027} & \textbf{0.0040} & \textbf{0.0045} & \textbf{0.0037} \\
LAW(v+r)     & Real    & \textbf{0.26}  & \textbf{0.39}  & \textbf{0.51}  & \textbf{0.39}  & 0.0027 & 0.0045 & 0.0057 & 0.0043 \\
\hline
\end{tabular}%
}
\label{tab:FreeWorldExperiments}
\end{table*}

\section{Experiments}

\subsection{Implementation Details}
We conducted experiments on the FreeWorld dataset, which consists of 309 unstructured road scenes. The dataset comprises approximately 1.5 million 3D bounding boxes spanning 16 object categories. Scene images are captured using six cameras that collectively provide a 360° horizontal field of view, with frames annotated at 2 Hz in the real-world portion and 1 Hz in the virtual portion. Following prior work \cite{jiang2023vad}, we evaluate planning performance using displacement error (DE), composition-based average precision (AP) for divider detection, and collision rate (CR) across different models and dataset subsets.




We conducted fine-tuning experiments on both virtual and real datasets to evaluate our models. Three dataset configurations were considered: virtual-only (V), real-only (R), and combined virtual+real (V+R).

For the \textbf{LAW (perception-free)} model:
\begin{itemize}
    \item \textbf{V:} fine-tuned for 3 epochs on the virtual dataset.
    \item \textbf{R:} fine-tuned for 1 epoch on the real dataset.
    \item \textbf{V+R:} first fine-tuned for 3 epochs on the virtual dataset, followed by 1 epoch on the real dataset.
\end{itemize}

For the \textbf{VAD-Base} model, fine-tuning was performed in two stages:
\begin{itemize}
    \item \textbf{V:} Stage 1, fine-tuned for 3 epochs on the virtual dataset; Stage 2, fine-tuned for 1 epoch.
    \item \textbf{R:} Stage 1, fine-tuned for 1 epoch on the real dataset; Stage 2, fine-tuned for 1 epoch.
    \item \textbf{V+R:} Stage 1, fine-tuned for 3 epochs on the virtual dataset and 1 epoch on the real dataset; Stage 2, fine-tuned for 1 epoch on each dataset.
\end{itemize}

\subsection{Results And Analysis}

As shown in Table~\ref{tab:FreeWorldExperiments}, the fine-tuned LAW model consistently outperforms the VAD model. This indicates that LAW not only achieves state-of-the-art (SOTA) performance in autonomous driving scenarios, but also demonstrates that implicit representation is generally more effective than manually designed rules in unstructured environments. This observation is well-justified: unstructured scenarios are inherently more complex and require a deeper and more comprehensive understanding of the environment, whereas rule-based approaches may overlook critical information, limiting their effectiveness. Consequently, LAW exhibits superior performance across both the virtual and real parts of our dataset.

It is also worth noting that the original LAW model exhibits substantially larger L2 errors, even exceeding those of the original VAD. Its collision rate on real data is therefore not meaningful, as the excessively large L2 errors render the model’s behavior entirely inconsistent with the task. After fine-tuning, however, LAW achieves performance far superior to VAD. This phenomenon can be explained by the significant discrepancies between the implicit features of structured and unstructured environments, which lead LAW to display unreasonable behaviors when directly applied to unstructured scenarios.

In addition, models trained with additional virtual data (that is, LAW(v) and LAW(v + r) on virtual testing data, and LAW(v + r) on real testing data) generally achieve lower L2 errors. Moreover, under the same testing dataset, models trained solely on virtual data and those trained solely on real data exhibit very similar performance in both L2 errors and collision rates. This suggests that our collected virtual and real data ----or the underlying scenes ----are perceived by the model as belonging to the same distribution. Furthermore, models fine-tuned with real data achieve lower collision rates, likely because the base models were originally trained on datasets such as nuScenes, which are derived from real-world environments. Consequently, fine-tuning with real data proves more effective for adapting to target and motion prediction in unstructured environments, whereas training exclusively on virtual data provides comparatively limited benefits.

\subsection{Qualitative Results}
\begin{figure}[H]
    \centering
    \includegraphics[width=1\linewidth]{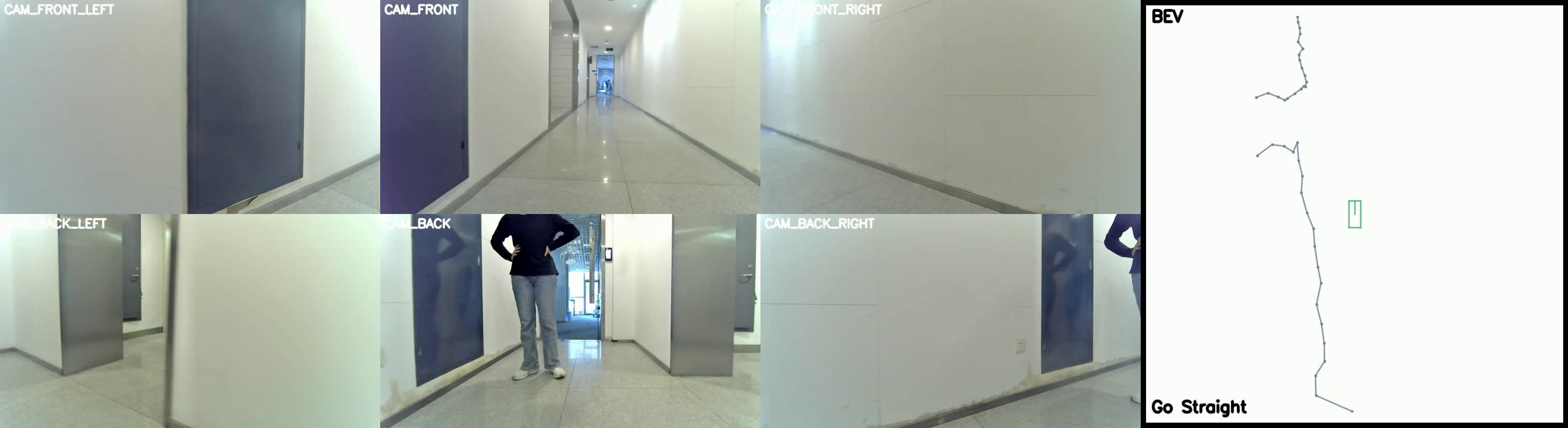}
    \caption{Qualitative results of VAD on the FreeWorld dataset.}
    \label{fig:VAD_qualitative_results_compressed}
\end{figure}

We show three vectorized scene learning and planning results of VAD(r) in Fig.~\ref{fig:qualitative_analysis_compressed} and VAD results in Fig.\ref{fig:VAD_qualitative_results_compressed}. For a better understanding of the scene, we also provide raw surrounding camera images 

\section{Conclusions and Future Work}

\subsection{Conclusions}
In this work, we present Bench2FreeAD, a comprehensive benchmark for end-to-end (E2E) robot navigation in unstructured environments. Our main contributions include the development of the FreeWorld dataset, which consists of both real-world and synthetic data collected through carefully designed pipelines. The dataset provides rich annotations for static and dynamic objects, enabling robust training and evaluation of E2E navigation models.

We fine-tuned two state-of-the-art E2E autonomous driving models, VAD and LAW, on our dataset and demonstrated significant improvements in navigation performance within unstructured scenarios. Experimental results show that the LAW model, which leverages implicit representations, consistently outperforms the VAD model, especially in complex and ambiguous environments. Furthermore, our analysis indicates that virtual and real data are perceived as similar by the models, and fine-tuning with real data further enhances collision avoidance.

By establishing this benchmark and dataset, we provide a solid foundation for future research in E2E navigation for logistics and service robots operating in challenging, unstructured environments.

\subsection{Future Work}
While our benchmark and dataset address several key challenges in unstructured robot navigation, there remain important directions for future research:

\begin{itemize}
    \item Dataset Expansion and Diversity: We plan to further expand the FreeWorld dataset to include more diverse environments, such as industrial sites, crowded public spaces, and outdoor natural terrains. Increasing the variety and scale of data will help improve model generalization and robustness.
    
    \item Real-world Deployment and Closed-loop Evaluation: We aim to deploy the fine-tuned models on physical robots for closed-loop navigation tasks, enabling evaluation of real-time performance, safety, and adaptability in dynamic, unpredictable environments.
    
    \item Advanced Method Paradigms: Future work will also explore the development of Vision-Language-Action (VLA) models, which tightly couple perception, reasoning, and decision-making. By leveraging language-guided representations together with visual and action signals, VLA-based approaches could significantly enhance the adaptability, interpretability, and efficiency of E2E navigation models in unstructured environments.
    
    \item Benchmark Extension: We will continue to refine and extend our benchmark metrics to cover additional aspects of navigation, such as social compliance, energy efficiency, and long-term autonomy.
\end{itemize}

\section{Appendix}
\subsection{Additional Qualitative Results}

\subsection{Maps}

\begin{figure}[H]
    \centering
    \includegraphics[width=4cm]{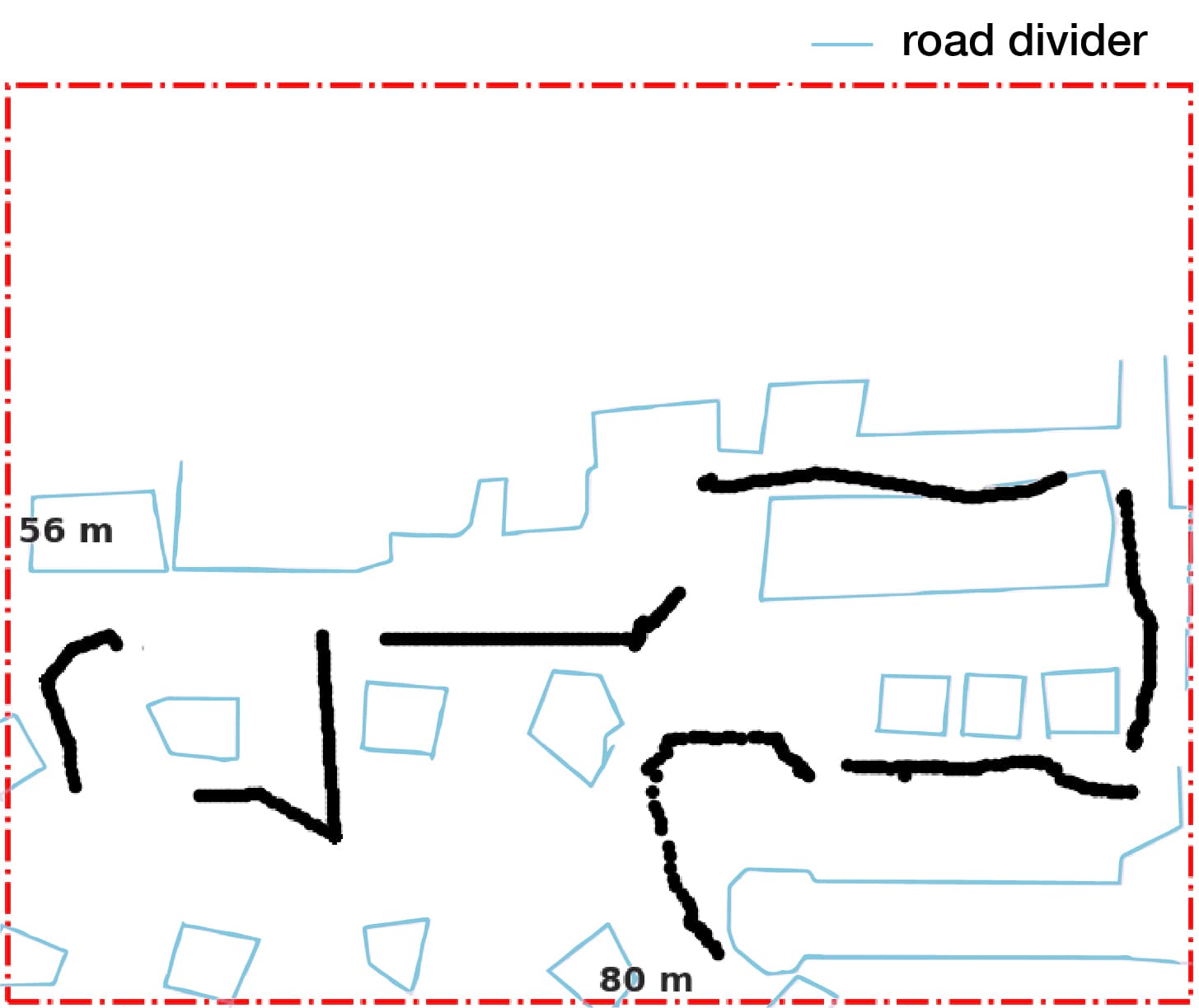}
    \caption{Visualization of ego poses in the \textit{RealWorld Map 1}}
    \label{fig:ego2img_AIR_F1_compressed1}
\end{figure}

\begin{figure}[H]
    \centering
    \includegraphics[width=4cm]
    {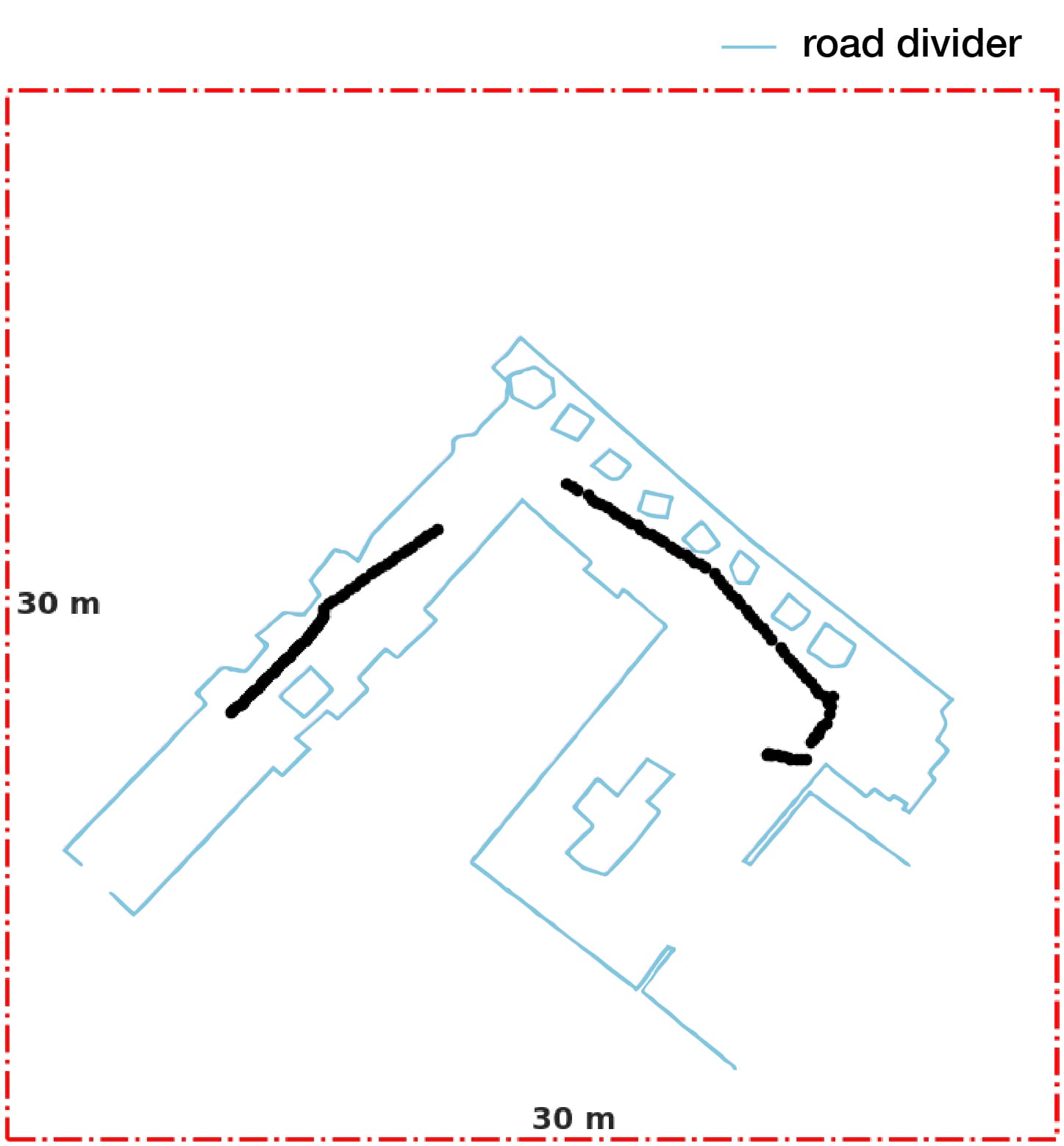}
    \caption{Visualization of ego poses in the \textit{RealWorld Map 2}}
    \label{fig:ego2img_AIR_G_compressed}
\end{figure}

\begin{figure}[H]
    \centering
    \includegraphics[width=4cm]{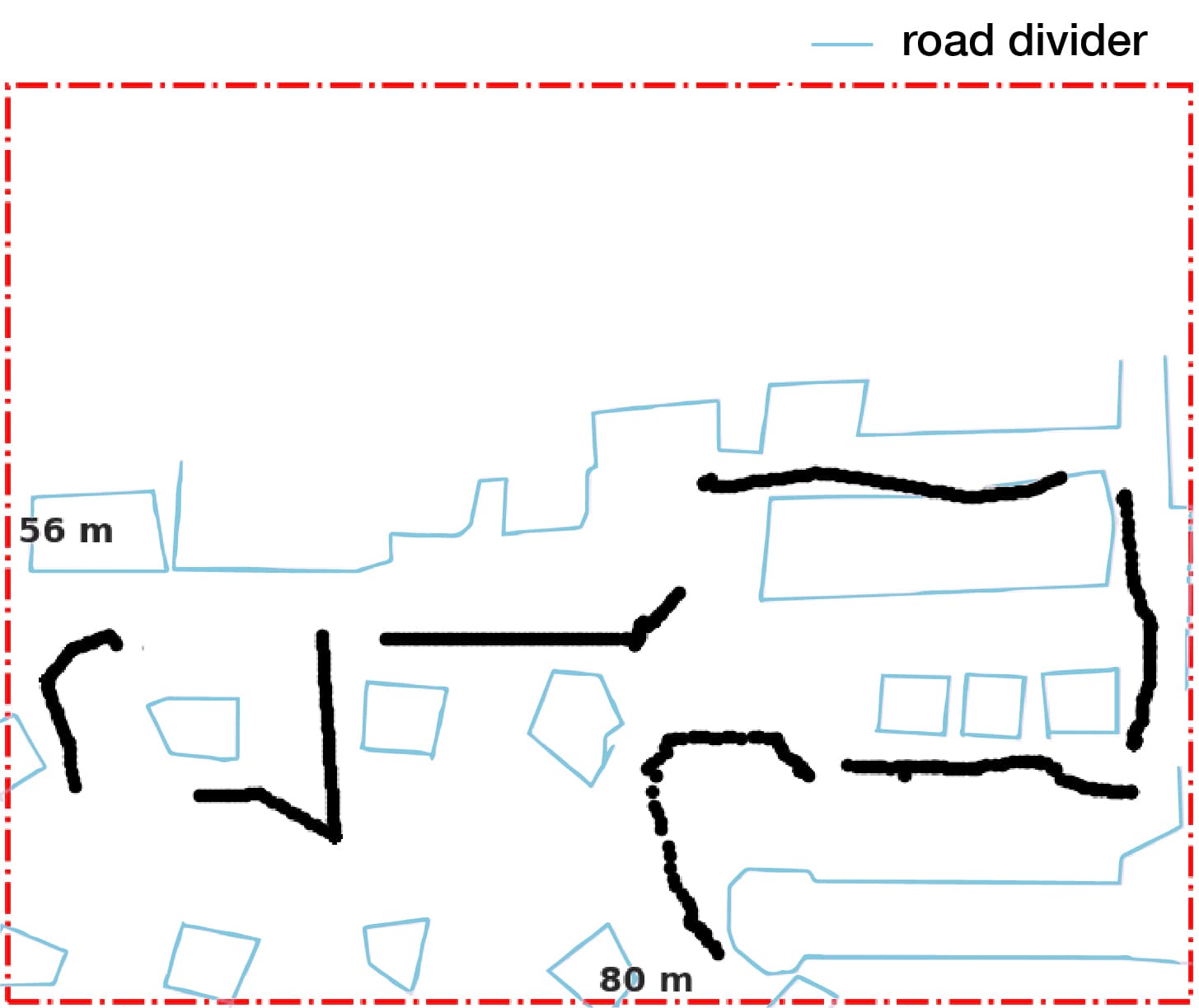}
    \caption{Visualization of ego poses in the \textit{RealWorld Map 3}}
    \label{fig:ego2img_AIR_F1_compressed}
\end{figure}

\bibliographystyle{IEEEtran}
\bibliography{IEEEfull,root_not_Anonymous}

\end{document}